\documentclass[conference]{IEEEtran}

\usepackage{graphicx}
\usepackage{amsmath,amsfonts,amssymb}
\usepackage{booktabs}
\usepackage{algorithm}
\usepackage{algorithmic}
\usepackage{cite}

\IEEEoverridecommandlockouts

\title{COMET-SG1: A Stability-Oriented Behavior-Space Autoregressive Regressor for Edge AI}

\author{
\IEEEauthorblockN{Shakhyar Gogoi}
\IEEEauthorblockA{
Jorhat Engineering College (Civil Eng.)\\
Indian Institute of Technology Guwahati (B.Sc. Hons, Artificial Intelligence)\\
Email: g.shakhyar@op.iitg.ac.in\\
Alternate: shakhyargogoi7@gmail.com
}
}

\begin{document}
\maketitle

\begin{abstract}
Edge AI systems deployed on microcontroller-class hardware impose strict constraints on memory, computation, determinism, and long-horizon stability. Conventional deep sequence models prioritize short-horizon accuracy and frequently exhibit unstable behavior when rolled out autoregressively, making them unsuitable for embedded settings. We present \emph{COMET-SG1}, a stability-oriented autoregressive regression framework explicitly designed around edge constraints. COMET-SG1 encodes short- and long-term temporal context into a compact behavior space and advances predictions through memory-anchored state transitions rather than recurrence or attention. The model relies on linear projections and L1 distance operations, avoids inference-time matrix multiplications, and exhibits bounded long-horizon behavior by construction. Experiments on non-stationary synthetic trading-like time series demonstrate that while feedforward and recurrent baselines achieve competitive short-horizon accuracy, they suffer from monotonic drift or oscillatory instability under autoregressive rollout. In contrast, COMET-SG1 maintains consistent bounded deviation across multiple random seeds with a compact footprint suitable for edge deployment. This work targets edge and embedded deployment scenarios such as Arduino- and microcontroller-class systems, where deterministic execution and long-horizon stability are critical.

\end{abstract}


\section{Introduction}

Autoregressive time-series prediction is a core primitive in many edge AI applications including sensing, monitoring, and control. Unlike cloud-based inference, edge deployments must operate under strict constraints on memory, computation, power consumption, and determinism. While deep sequence models such as LSTMs and Transformers perform well on offline forecasting benchmarks, they are poorly matched to microcontroller-class hardware and frequently exhibit unstable long-horizon behavior.

A central difficulty in autoregressive inference is error accumulation: small one-step prediction errors compound over time, leading to divergence or collapse. Models trained purely for pointwise accuracy offer no explicit mechanism to control this phenomenon. In embedded settings, where frequent re-anchoring to ground truth may be infeasible, stability under long-horizon rollout is often more important than marginal gains in short-horizon accuracy.

This work introduces \emph{COMET-SG1}, a behavior-space autoregressive regressor designed explicitly around edge constraints. Rather than modeling temporal evolution through recurrence, COMET-SG1 represents local dynamics in a compact behavior space and advances predictions using memory-anchored transitions derived from historical trajectories. This design prioritizes bounded behavior, predictability, and deployability.

\section{Related Work}

Classical autoregressive and linear regression models are computationally lightweight but struggle with non-stationary dynamics and regime changes. Memory-based approaches such as k-nearest neighbors (kNN) adapt locally to data but are sensitive to noise and often unstable under autoregressive rollout.

Deep learning approaches including multilayer perceptrons (MLPs) and recurrent neural networks (LSTMs) achieve strong short-horizon accuracy but frequently exhibit monotonic drift or oscillatory instability during long-horizon rollout. Moreover, their reliance on dense matrix multiplications and recurrent computation limits their suitability for embedded systems. Recent TinyML research emphasizes compression and quantization, but stability under autoregressive inference remains largely unaddressed.

\section{Methodology}

\subsection{Problem Definition}

Let $\{x_t\}_{t=1}^{\infty}$, $x_t \in \mathbb{R}$, denote a univariate time series sampled at uniform intervals. Given a fixed-length history window
\[
\mathbf{x}_t^{(L)} = [x_{t-L+1}, x_{t-L+2}, \dots, x_t],
\]
the task is to predict the next value $\hat{x}_{t+1}$. During deployment, predictions are generated autoregressively, meaning that each predicted value is appended to the history window and subsequently used as input for future predictions. This recursive usage makes the system sensitive to error accumulation, requiring explicit mechanisms to control long-horizon behavior.

\subsection{Multi-Scale Behavior-Space Encoding}

COMET-SG1 represents recent temporal behavior using linear projections applied to multiple window lengths. Specifically, three encoders are defined:
\begin{align}
\mathbf{z}_s &= E_s(\mathbf{x}_t^{(12)}) \in \mathbb{R}^D, \\
\mathbf{z}_m &= E_m(\mathbf{x}_t^{(24)}) \in \mathbb{R}^D, \\
\mathbf{z}_l &= E_l(\mathbf{x}_t^{(60)}) \in \mathbb{R}^D.
\end{align}
Each encoder $E_\ast : \mathbb{R}^{L_\ast} \rightarrow \mathbb{R}^D$ is a bias-free linear map. No recurrence, gating, or nonlinear activation is used. The encoders operate directly on raw values rather than differences or normalized inputs.

The resulting vectors $\mathbf{z}_s$, $\mathbf{z}_m$, and $\mathbf{z}_l$ provide compact summaries of short-, medium-, and long-term temporal behavior, respectively. Importantly, these representations are recomputed independently at each time step and do not depend on previous internal states.

\subsection{Internal Behavior State}

In addition to the instantaneous encodings, COMET-SG1 maintains an internal behavior state
\[
\mathbf{z}_t \in \mathbb{R}^D,
\]
initialized as $\mathbf{z}_0 = \mathbf{0}$. This state is not directly derived from the input window; instead, it accumulates inferred behavioral transitions over time. The internal state provides temporal continuity across autoregressive steps and acts as a latent trajectory variable.

\subsection{Memory Construction During Training}

During offline training, COMET-SG1 constructs a memory of observed transitions from ground-truth data. At each time index $i$, the following quantities are computed:
\begin{itemize}
\item Behavior encodings $(\mathbf{z}_{s,i}, \mathbf{z}_{m,i}, \mathbf{z}_{l,i})$
\item Output increment $\Delta x_i = x_{i+1} - x_i$
\item Behavior-space increment $\Delta \mathbf{z}_i = \mathbf{z}_{s,i} - \mathbf{z}_{s,i-1}$
\end{itemize}

Each memory entry is stored as
\[
\mathcal{M}_i =
(\mathbf{z}_{s,i}, \mathbf{z}_{m,i}, \mathbf{z}_{l,i},
\Delta \mathbf{z}_i, \Delta x_i).
\]

The full memory set $\mathcal{M} = \{\mathcal{M}_1, \dots, \mathcal{M}_N\}$ therefore represents a collection of local transition examples, each describing how a particular behavior-space configuration evolved in the data.

\subsection{Distance Metric and Neighborhood Retrieval}

At inference time, given current encodings $(\mathbf{z}_s, \mathbf{z}_m, \mathbf{z}_l)$, similarity to memory entries is measured using a weighted L1 distance:
\[
d_i =
w_s \|\mathbf{z}_s - \mathbf{z}_{s,i}\|_1 +
w_m \|\mathbf{z}_m - \mathbf{z}_{m,i}\|_1 +
w_l \|\mathbf{z}_l - \mathbf{z}_{l,i}\|_1,
\]
where $w_s, w_m, w_l > 0$ are learned scalar coefficients. This metric allows the model to adaptively emphasize different temporal scales during retrieval.

The $K$ memory entries with the smallest distances are selected to form the neighborhood $\mathcal{N}_K$.

\subsection{Memory-Anchored Transition Aggregation}

To aggregate information from the selected neighborhood, a soft weighting scheme is applied. Raw attention weights are computed as
\[
\alpha_i = \frac{\exp(-\gamma d_i)}{\sum_{j \in \mathcal{N}_K} \exp(-\gamma d_j)},
\]
where $\gamma > 0$ controls the sharpness of the weighting. To prevent weight collapse and encourage smoother aggregation, the weights are mixed with a uniform component:
\[
\alpha_i \leftarrow 0.7 \alpha_i + \frac{0.3}{K}.
\]

Using these weights, the memory-anchored transitions are estimated as
\begin{align}
\Delta \mathbf{z}_{\text{mem}} &= \sum_{i \in \mathcal{N}_K} \alpha_i \Delta \mathbf{z}_i, \\
\Delta x_{\text{mem}} &= \sum_{i \in \mathcal{N}_K} \alpha_i \Delta x_i.
\end{align}

These quantities represent the expected behavior-space and output increments inferred from similar past situations.

\subsection{Learned Correction Term}

While memory aggregation captures locally observed transitions, it may be insufficient in regions sparsely represented in memory. To address this, COMET-SG1 includes a learned correction term:
\[
\Delta \mathbf{z}_{\text{learned}} =
W_f [\mathbf{z}_t; \mathbf{z}_s; \mathbf{z}_m; \mathbf{z}_l],
\]
where $W_f$ is a linear projection and $[\cdot;\cdot]$ denotes vector concatenation. This term allows the model to interpolate and generalize beyond stored examples while remaining linear and lightweight.

\subsection{State and Output Update}

The internal behavior state is updated as
\[
\mathbf{z}_{t+1} =
\mathbf{z}_t +
\Delta \mathbf{z}_{\text{mem}} +
\Delta \mathbf{z}_{\text{learned}}.
\]

The predicted next value is obtained by applying the inferred output increment:
\[
\hat{x}_{t+1} = x_t + \Delta x_{\text{mem}}.
\]

This update structure ensures that long-horizon behavior emerges from accumulated, memory-anchored transitions rather than repeated application of a single-step predictor.

\subsection{Training Objective}

The model is trained using a one-step Huber loss:
\[
\mathcal{L} = \text{Huber}(\hat{x}_{t+1}, x_{t+1}),
\]
without multi-step or rollout loss terms. Stability emerges from the model structure rather than the objective.

\begin{algorithm}[t]
\caption{COMET-SG1 Autoregressive Inference}
\label{alg:comet}
\begin{algorithmic}[1]
\STATE Encode history into $(\mathbf{z}_s,\mathbf{z}_m,\mathbf{z}_l)$
\STATE Compute distances to memory entries
\STATE Select top-$K$ neighbors
\STATE Aggregate memory transitions
\STATE Apply learned correction
\STATE Update state $\mathbf{z}_{t+1}$ and output $\hat{x}_{t+1}$
\end{algorithmic}
\end{algorithm}

\section{Experimental Setup}

Synthetic trading-like time series exhibiting non-stationarity and regime changes are used. Baselines include kNN regression, a feedforward MLP, and an LSTM. Evaluation focuses on short-horizon accuracy and long-horizon autoregressive stability across multiple random seeds.

\section{Embedded Feasibility Analysis}

COMET-SG1 uses approximately 2.9~KB of parameters, with memory dominated by stored behavior entries ($\sim$345~KB). Inference consists of linear projections, L1 distance computations, weighted summations, and simple state updates. The absence of recurrence, attention, and dynamic allocation makes the approach compatible with fixed-point microcontroller implementations. Python timing reflects prototype overhead rather than deployment cost.

\subsection{Edge and Microcontroller Suitability}
The design of COMET-SG1 aligns closely with the constraints of microcontroller-class edge platforms such as Arduino- and Cortex-M–based systems. In contrast to recurrent or attention-based models, inference consists exclusively of linear projections, L1 distance computations, scalar weighting, and vector accumulation. These operations map directly to fixed-point arithmetic and do not require dynamic memory allocation, recursion, or stack-intensive control flow.

From an embedded systems perspective, the absence of recurrence eliminates the need for maintaining hidden-state tensors across time steps, while the lack of attention mechanisms avoids quadratic memory access patterns. All model parameters are static, and memory access during inference is sequential and deterministic. Such properties are critical for predictable execution latency, bounded RAM usage, and energy efficiency on ultra-low-power devices.

Importantly, the behavior-space formulation decouples model complexity from temporal horizon. Increasing the prediction horizon does not increase model depth, parameter count, or inference graph complexity, making COMET-SG1 particularly well suited for long-horizon deployment scenarios on constrained hardware. This contrasts with neural sequence models, where longer horizons exacerbate instability and compound computational cost.

While the experimental evaluation is conducted using a Python prototype, the underlying operations required by COMET-SG1 are compatible with existing TinyML toolchains and microcontroller inference frameworks. As such, the proposed model represents a practical step toward stability-oriented time-series regression for edge and embedded AI systems.
\subsection{Inference Timing Comparison}
Table~\ref{tab:cost} reports prototype-level inference time per prediction step and model footprint measured using a Python implementation. Absolute timing values reflect interpreter and memory-scan overhead and should not be interpreted as deployment latency. Instead, the comparison highlights relative computational structure and memory footprint across models.

\begin{table}[t]
\centering
\caption{Inference Cost and Model Footprint (Prototype Measurement)}
\label{tab:cost}
\begin{tabular}{lccc}
\toprule
Model & Time / Step (s) & Params (KB) & Memory (KB) \\
\midrule
kNN   & 0.00172 & --    & -- \\
MLP   & 0.00103 & 31.8  & -- \\
LSTM  & 0.00145 & 17.6  & -- \\
COMET-SG1 & 0.50 & \textbf{2.9} & 345 \\
\bottomrule
\end{tabular}
\end{table}

While COMET-SG1 exhibits higher prototype inference time due to full memory scans, its parameter footprint is substantially smaller than neural baselines. Importantly, inference consists of linear projections, L1 distance computations, and weighted summations, all of which map cleanly to fixed-point arithmetic. Memory access patterns are sequential and static, making the approach suitable for microcontroller-class hardware despite prototype overhead.

\section{Results}

\subsection{Short-Horizon Accuracy}

Table~\ref{tab:mae} reports mean absolute error (MAE) for one-step and five-step prediction.

\begin{table}[t]
\centering
\caption{Short-Horizon Prediction Accuracy (MAE)}
\label{tab:mae}
\begin{tabular}{lcc}
\toprule
Model & 1-step MAE & 5-step MAE \\
\midrule
kNN & 0.0111 & 0.0155 \\
MLP & 0.0107 & 0.0167 \\
LSTM & 0.0347 & 0.0808 \\
COMET-SG1 & \textbf{0.0061} & 0.0177 \\
\bottomrule
\end{tabular}
\end{table}

\subsection{Long-Horizon Rollout Stability}
\begin{figure*}[t]
\centering
\includegraphics[width=0.48\textwidth]{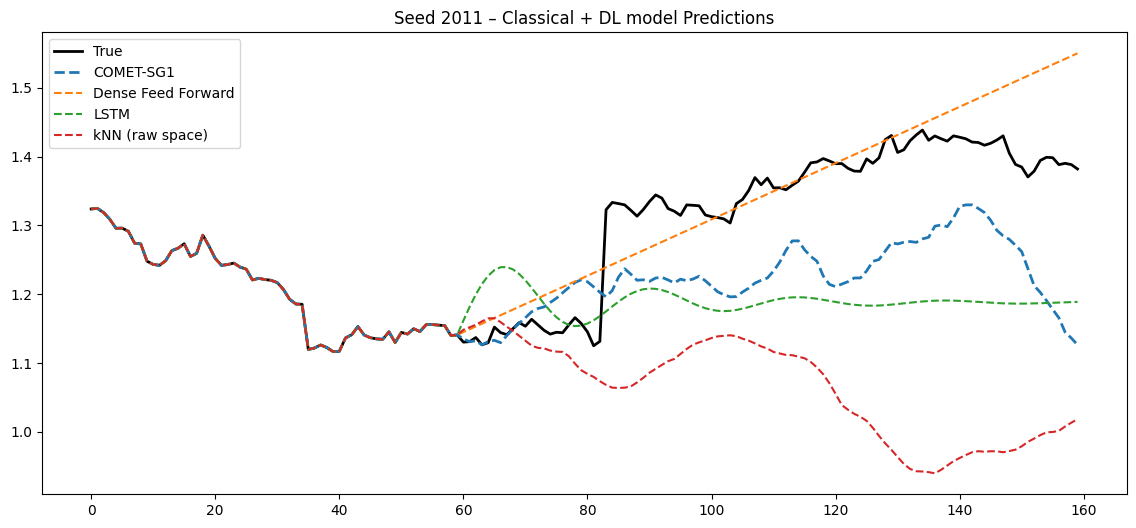}
\includegraphics[width=0.48\textwidth]{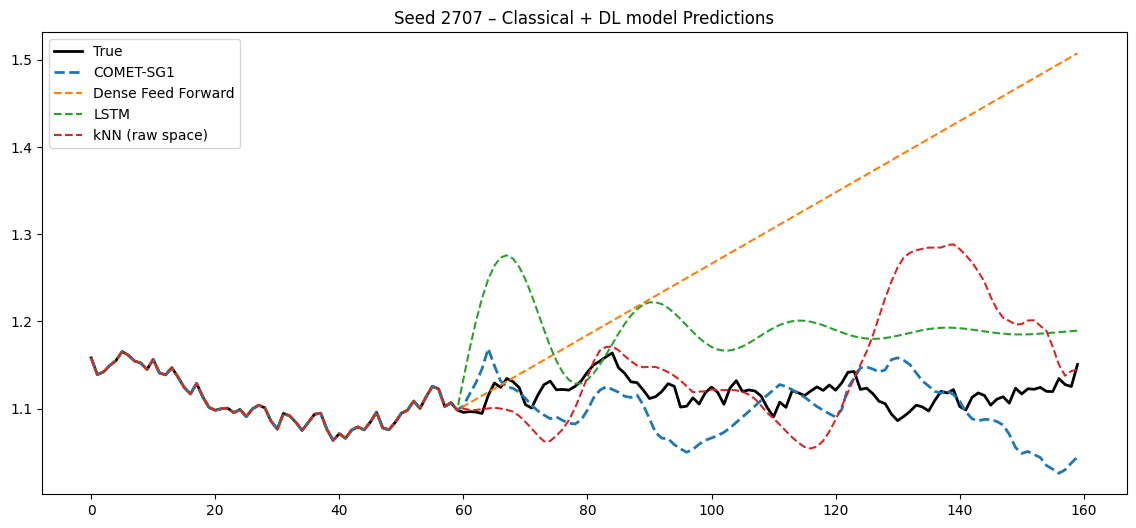}\\
\includegraphics[width=0.48\textwidth]{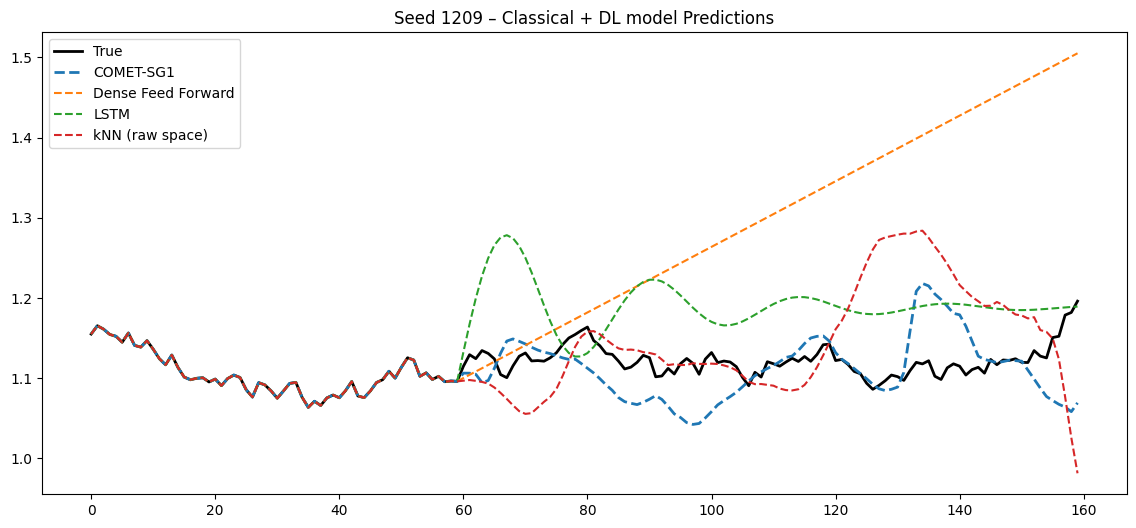}
\includegraphics[width=0.48\textwidth]{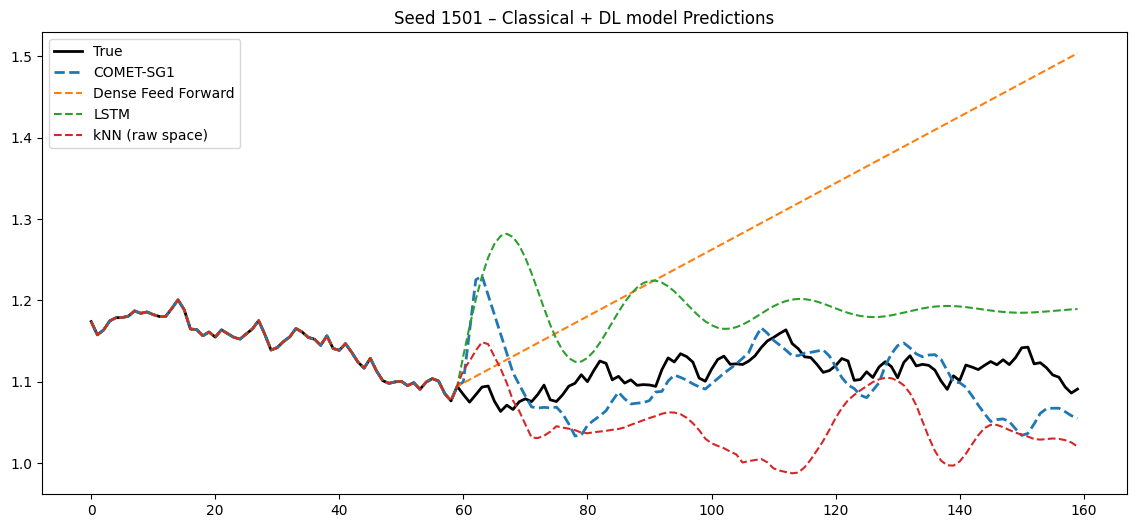}

\caption{Autoregressive rollout comparisons across four random seeds.}
\label{fig:rollout}
\end{figure*}

Figure~\ref{fig:rollout} illustrates long-horizon autoregressive rollouts for four representative random seeds. While feedforward (MLP) and recurrent (LSTM) baselines exhibit monotonic drift or oscillatory instability when predictions are recursively fed back as input, COMET-SG1 maintains bounded deviation across all seeds. This behavior indicates that anchoring updates to behavior-space memory transitions effectively limits error accumulation during long-horizon inference.

\subsection{Drift Analysis}
\begin{figure}[t]
\centering

\includegraphics[width=\linewidth]{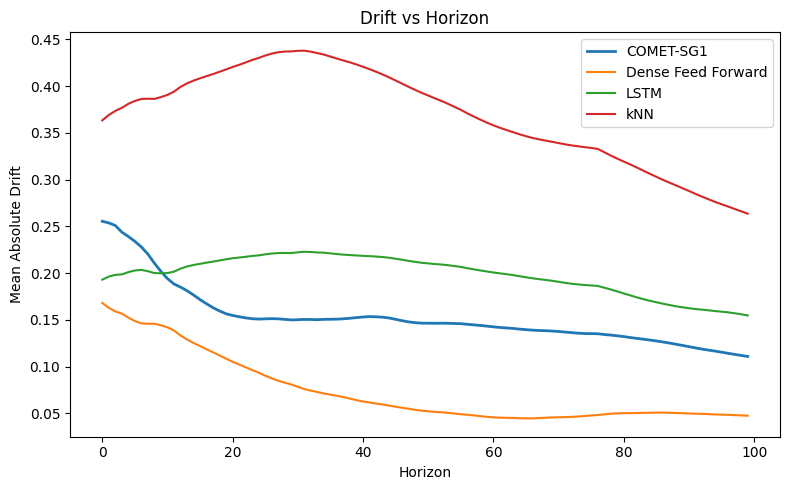}
\caption{Mean absolute drift as a function of prediction horizon.}
\label{fig:drift}
\end{figure}

To quantitatively evaluate long-horizon stability, Fig.~\ref{fig:drift} reports the mean absolute drift as a function of prediction horizon. Unlike kNN and LSTM baselines, which exhibit increasing or plateaued drift, COMET-SG1 shows bounded and gradually decreasing drift with horizon, confirming the qualitative rollout behavior observed in Fig.~\ref{fig:rollout}.

\section{Discussion}

The results presented in this work highlight the importance of stability-oriented design for autoregressive inference under edge constraints. While many prior approaches emphasize short-horizon accuracy, the experiments demonstrate that such objectives alone are insufficient to guarantee reliable long-horizon behavior when predictions are recursively fed back as input. COMET-SG1 addresses this gap by anchoring updates to behavior-space memory transitions, resulting in bounded deviation across diverse regimes and random seeds.

An important characteristic of the proposed approach is the clear separation between representational capacity and inference structure. Rather than relying on increasingly complex architectures, COMET-SG1 leverages simple linear encoders and distance-based retrieval to capture local dynamics. This design choice enables predictable behavior under rollout while remaining compatible with deterministic and resource-constrained execution environments.

The current formulation also provides a flexible foundation for future extensions. For example, alternative memory indexing strategies, approximate neighbor retrieval, or adaptive memory management policies can be integrated without altering the core inference mechanism. Similarly, the behavior-space representation can be enriched with additional descriptors when higher-dimensional signals are available, while preserving the same stability-oriented update structure.

\section{Conclusion}

We presented COMET-SG1, a stability-oriented autoregressive regressor explicitly designed around edge AI constraints. By prioritizing bounded long-horizon behavior, the model achieves predictable rollout while maintaining a compact footprint suitable for embedded systems.

\bibliographystyle{IEEEtran}

\begin{thebibliography}{99}

\bibitem{tinyml}
P. Warden and D. Situnayake, \emph{TinyML: Machine Learning with TensorFlow Lite on Arduino and Ultra-Low-Power Microcontrollers}, O’Reilly, 2020.

\bibitem{lstm}
S. Hochreiter and J. Schmidhuber, ``Long Short-Term Memory,'' \emph{Neural Computation}, vol. 9, no. 8, pp. 1735--1780, 1997.

\bibitem{gru}
K. Cho \emph{et al.}, ``Learning Phrase Representations using RNN Encoder--Decoder for Statistical Machine Translation,'' \emph{EMNLP}, 2014.

\bibitem{knn}
T. M. Cover and P. E. Hart, ``Nearest Neighbor Pattern Classification,'' \emph{IEEE Trans. Information Theory}, vol. 13, no. 1, pp. 21--27, 1967.

\bibitem{arima}
G. E. P. Box and G. M. Jenkins, \emph{Time Series Analysis: Forecasting and Control}, Holden-Day, 1970.

\bibitem{attention}
A. Vaswani \emph{et al.}, ``Attention Is All You Need,'' \emph{NeurIPS}, 2017.

\bibitem{tcn}
S. Bai, J. Z. Kolter, and V. Koltun, ``An Empirical Evaluation of Generic Convolutional and Recurrent Networks for Sequence Modeling,'' \emph{arXiv:1803.01271}, 2018.

\bibitem{nbeats}
B. Oreshkin, D. Carpov, N. Chapados, and Y. Bengio, ``N-BEATS: Neural Basis Expansion Analysis for Interpretable Time Series Forecasting,'' \emph{ICLR}, 2020.

\bibitem{deepar}
D. Salinas \emph{et al.}, ``DeepAR: Probabilistic Forecasting with Autoregressive Recurrent Networks,'' \emph{IJCAI}, 2020.

\bibitem{kalman}
R. E. Kalman, ``A New Approach to Linear Filtering and Prediction Problems,'' \emph{ASME Journal of Basic Engineering}, 1960.

\bibitem{reservoir}
H. Jaeger, ``The Echo State Approach to Analysing and Training Recurrent Neural Networks,'' GMD Technical Report, 2001.

\bibitem{memorynet}
J. Weston, S. Chopra, and A. Bordes, ``Memory Networks,'' \emph{ICLR}, 2015.

\bibitem{edgeai}
S. Han, H. Mao, and W. J. Dally, ``Deep Compression,'' \emph{ICLR}, 2016.

\bibitem{quant}
B. Jacob \emph{et al.}, ``Quantization and Training of Neural Networks for Efficient Integer-Arithmetic-Only Inference,'' \emph{CVPR}, 2018.

\bibitem{conceptdrift}
J. Gama \emph{et al.}, ``A Survey on Concept Drift Adaptation,'' \emph{ACM Computing Surveys}, 2014.

\bibitem{ts_survey}
G. Lai \emph{et al.}, ``Modeling Long- and Short-Term Temporal Patterns with Deep Neural Networks,'' \emph{SIGIR}, 2018.

\end{thebibliography}

\end{document}